\documentclass[letterpaper]{article} 
\usepackage[preprint]{aaai2027}  
\usepackage[hyphens]{url}  
\usepackage{graphicx} 
\usepackage{natbib}  
\usepackage{caption} 
\usepackage{algorithm}
\usepackage{algorithmic}

\usepackage{newfloat}
\usepackage{listings}
\DeclareCaptionStyle{ruled}{labelfont=normalfont,labelsep=colon,strut=off} 
\floatstyle{ruled}
\newfloat{listing}{tb}{lst}{}
\floatname{listing}{Listing}

\usepackage{booktabs}
\usepackage{colortbl}
\usepackage{multirow}
\usepackage{microtype}

\definecolor{oakblue}{RGB}{232,241,250}
\definecolor{codebg}{RGB}{238,238,238}
\definecolor{diffaddbg}{RGB}{226,244,230}
\lstdefinestyle{csblock}{
  backgroundcolor=\color{codebg},
  basicstyle=\footnotesize\ttfamily,
  frame=single,
  framesep=4pt,
  xleftmargin=0pt,
  numbers=none,
  breaklines=true,
  showstringspaces=false,
  tabsize=2,
  aboveskip=6pt,
  belowskip=6pt
}

\title{Toward Effective and Reliable LLM Agents via Dynamic Ontology}
\author{
    Xiaohui Zhang\textsuperscript{\rm 1},
    Zequn Sun\textsuperscript{\rm 1},
    Chengyuan Yang\textsuperscript{\rm 1},
    Yuanning Cui\textsuperscript{\rm 2},
    Lingbing Guo\textsuperscript{\rm 3},
    Wei Hu\textsuperscript{\rm 1,\rm 4}
}
\affiliations{
    \textsuperscript{\rm 1}State Key Laboratory for Novel Software Technology, Nanjing University, China\\
    \textsuperscript{\rm 2}School of Computer Science, Nanjing University of Information Science and Technology, China\\
    \textsuperscript{\rm 3}School of Intelligence Science and Technology, Nanjing University, China\\
    \textsuperscript{\rm 4}National Institute of Healthcare Data Science, Nanjing University, China\\
    \{xhzhang, cyyang\}.nju@gmail.com, \{sunzq, lbguo, whu\}@nju.edu.cn, yncui@nuist.edu.cn
}

\begin{document}

\maketitle

\begin{abstract}
Large language model (LLM) agents rely heavily on knowledge encoded in model parameters or presented as unstructured context. 
In domain-specific tasks, this leaves important semantic connections implicit. This often results in incomplete evidence use and brittle multi-step decisions. 
Ontologies offer a way to externalize domain concepts and relations as machine-interpretable structures, but constructing task-usable ontologies traditionally requires substantial effort from domain experts and is difficult to scale. 
Automatic construction is also challenging: an ontology that appears semantically plausible may not contain the relational structures needed for actual decision making. 
We present OaK, an ontology-as-a-kernel framework that dynamically constructs and refines task-oriented ontologies for LLM agents. 
Given task requirements and training data, OaK constructs an ontology and its knowledge graph, generates task-adaptation functions for graph reasoning, and uses judge feedback to iteratively refine both.
By making relevant concepts and relations explicit, the ontology grounds knowledge retrieval and multi-step decision making. 
We evaluate OaK on TravelPlanner, CRMArenaPro, and ToolQA.
Results show that OaK improves standard LLM agents, strengthens evidence grounding, and boosts the reliability of multi-step reasoning.
\end{abstract}

\section{Introduction}

Large language models (LLMs) have demonstrated strong capabilities in understanding and generating natural language, as well as in knowledge-intensive reasoning. LLMs address diverse tasks through instructions and in-context examples.
LLM agents extend these capabilities from response generation to goal-directed task execution by coupling an LLM with external components such as retrieval systems~\citep{DBLP:conf/acl/ZhuLSWH25}, tools and memory~\citep{DBLP:journals/corr/abs-2603-00026,DBLP:journals/corr/abs-2601-04463}.
A typical agent runs a loop that repeatedly interprets the goal and plans, then acts through tool calls or retrieval and observes the result until the task is done.
Recent agent architectures further incorporate mechanisms such as reflection and procedural memory, along with modular workflow optimization~\citep{reflexion2023,memp2025,aflow2025,agentsquare2025}. These developments make LLM agents a general framework for tasks that require command execution and multi-step reasoning beyond a single response.

Despite these advances, the central challenge is no longer only whether an agent can act, but whether its behavior remains controllable and trustworthy as execution becomes longer and more autonomous.
At each intermediate step, the agent decides what to retrieve and which tool to call with which arguments, so an error in any of these choices can propagate to later steps.
Consequently, final-answer accuracy alone reveals neither the supporting evidence nor the justification of tool calls. It also hides the influence of memory and the origin of execution failures~\citep{DBLP:journals/corr/abs-2606-04990}.
Recent studies of tool-using agents expose this gap: ToolEmu identifies realistic long-tail safety failures in high-stakes tool settings, while AgentDojo shows that untrusted tool outputs can manipulate agent behavior through prompt injection~\citep{toolem2024,agentdojo2024}.
Reflection, procedural memory, and workflow optimization can improve planning or reuse, but they generally leave the set of admissible concepts and action sequences implicit~\citep{selfrefine2023,reflexion2023,aflow2025,memp2025}.
Prompt instructions and tool descriptions therefore guide an agent's behavior without providing an enforceable contract for what it may execute or how its results should be checked.
Reliable agents must constrain actions during execution so that outputs connect to supporting evidence and faulty steps can be identified and fixed.

These requirements point to a missing layer between the LLM and the tools it controls: a task-oriented representation that makes both domain semantics and executable behavior explicit.
Ontologies provide a natural basis because they organize concepts and relations in a machine-interpretable form~\citep{gruber1993,knowledgegraphs2021}.
However, a conventional ontology is primarily descriptive: it specifies what exists in a domain, but not necessarily what an agent may do. It also leaves open how an operation should be parameterized and which conditions must hold before its result is accepted.
We therefore use ontology in an operational sense.
In our setting, requirements on data representation are encoded in schema declarations, whereas requirements on computation and workflow are enforced by the control flow of functions.
Such an ontology is not a passive knowledge description. It serves as a semantic and procedural contract that bounds what the agent may do and keeps execution open to inspection.

We propose OaK, a dynamic ontology-as-a-kernel framework for LLM agents.
Here, dynamicity refers to task-conditioned construction. 
For each task, OaK automatically constructs the schema and typed reasoning functions needed to solve it.
It then instantiates a corresponding knowledge graph from task data, and refines the schema and functions with training examples and downstream task feedback before freezing the resulting kernel for inference.
OaK packages the task interface into a kernel
\[
\mathcal{K}=(\mathcal{S},\mathcal{F}),
\]
where $\mathcal{S}$ is a task-oriented schema that defines the domain concepts available to the agent, together with their properties and relations.
The functions $\mathcal{F}$ define how these typed elements can be used through executable procedures for retrieval, filtering, traversal, projection, aggregation, and multi-step reasoning.
Given $\mathcal{S}$, OaK instantiates a schema-guided knowledge graph $\mathcal{G}$ from the task data. This graph provides the evidence on which the functions operate.
During inference, a ReAct agent interprets a query to select a function and bind its typed arguments, then uses the function to reason the final answer. The kernel mediates access to the available evidence and operations.
During construction, OaK evaluates these executions and uses task feedback to refine $\mathcal{S}$ and $\mathcal{F}$ before the resulting kernel is applied to unseen queries.

We evaluate OaK on TravelPlanner \cite{travelplanner2024}, CRMArenaPro \cite{crmarenaPro2025}, and ToolQA \cite{toolqa2023}.
These datasets differ in task setting and reasoning requirements.
Results and analyses show that OaK improves standard LLM agents and strengthens evidence grounding for multi-step reasoning.

Our contributions are summarized as follows:

\begin{itemize}
    \item We introduce \emph{ontology as a kernel} for LLM agents. It couples a task-oriented schema with typed reasoning functions and a schema-guided evidence graph.

    \item We develop an automated pipeline that constructs a verified schema, instantiates its knowledge graph, and compiles generic operators into executable domain functions.

    \item 
    We propose judge-driven refinement, which diagnoses and repairs schema and function failures across construction rounds using official task scores and execution trajectories.

    \item Experiments on TravelPlanner, CRMArenaPro, and ToolQA show consistent gains across two backbones, while ablations confirm the importance of function composition, the function module, and iterative refinement.
\end{itemize}

\section{Related Work}

\paragraph{LLM for agents.}
LLMs serve as general-purpose reasoning and decision-making components in agents. 
They enable agents to interpret goals and decompose tasks, adapting their behavior from intermediate observations.
ReAct established an influential paradigm that interleaves reasoning and acting within a single execution trajectory~\citep{react2023}.
Beyond reflection, MemP distills successful trajectories into reusable procedural memory, while ReCode represents planning and acting through recursively generated programs~\citep{memp2025,recode2025}.
AFlow and AgentSquare further automate agent design by searching over code-represented workflows or modular architectures composed of planning and memory components~\citep{aflow2025,agentsquare2025}.
These approaches progressively move agent control out of the model's parameters into feedback loops and reusable procedures.
They primarily improve how an agent organizes and adapts its control process, whereas OaK complements this line of work by structuring the task interface itself through an explicit domain schema and executable reasoning functions.

\paragraph{Reliable agents.}
Reliability in LLM agents extends beyond endpoint accuracy: it asks whether each step stays controlled and each decision remains grounded in evidence, and whether failures can be traced to a cause \cite{DBLP:journals/corr/abs-2606-04990}.
Feedback-based approaches such as Self-Refine and Reflexion use model-generated critique or execution outcomes to revise responses and future decisions~\citep{selfrefine2023,reflexion2023}.
ToolEmu and AgentDojo expose complementary risks by identifying realistic failures in high-stakes tool settings and evaluating indirect prompt injection through untrusted observations~\citep{toolem2024,agentdojo2024}.
AgentSpec addresses execution control by expressing safety requirements as structured rules, combining runtime triggers and predicates with enforcement mechanisms~\citep{agentspec2025}.
In parallel, knowledge editing aims to correct or update the facts stored in model parameters rather than externalizing them~\citep{knowledgeEditingNMI}.
These approaches provide feedback, policy enforcement, or execution records.
However, they typically treat task rules and allowable operations as fixed inputs or leave them distributed across prompts and implementation code.
As a result, the interface between task requirements and agent execution is difficult to inspect and adapt.
OaK instead makes this interface explicit as a task-specific kernel, coupling a schema with typed reasoning functions and using task scores and execution trajectories to refine both.

\paragraph{Graph for agents.}
Ontologies and knowledge graphs provide machine-interpretable structures for organizing domain concepts and their properties and relations~\citep{gruber1993,knowledgegraphs2021}.
Recent work integrates LLMs with knowledge graphs to support the structured acquisition and representation of knowledge, as well as reasoning over it~\citep{llmkg2023}.
GraphRAG and G-Retriever further use graph structure to retrieve globally relevant or relational evidence for generation and question answering~\citep{graphrag2024,gretriever2024}.
These approaches primarily treat the graph as an external knowledge and retrieval layer. 
This can improve evidence access but does not itself specify reusable task-level computations for an agent.
OaK instead constructs a task-oriented schema and a catalog of typed reasoning functions. It grounds their execution in a data-dependent knowledge graph.
It jointly refines the schema and functions using downstream task feedback.
In OaK, dynamicity therefore refers to constructing this task-specific schema and function catalog for the task at hand. It then instantiates the corresponding graph from task data and refines the kernel before freezing it for inference.
This design makes the graph not only a source of retrieved knowledge, but also the grounded substrate on which the agent's semantic and procedural interface operates.

\section{OaK Framework}

\begin{figure*}[t]
\centering
\includegraphics[width=\textwidth]{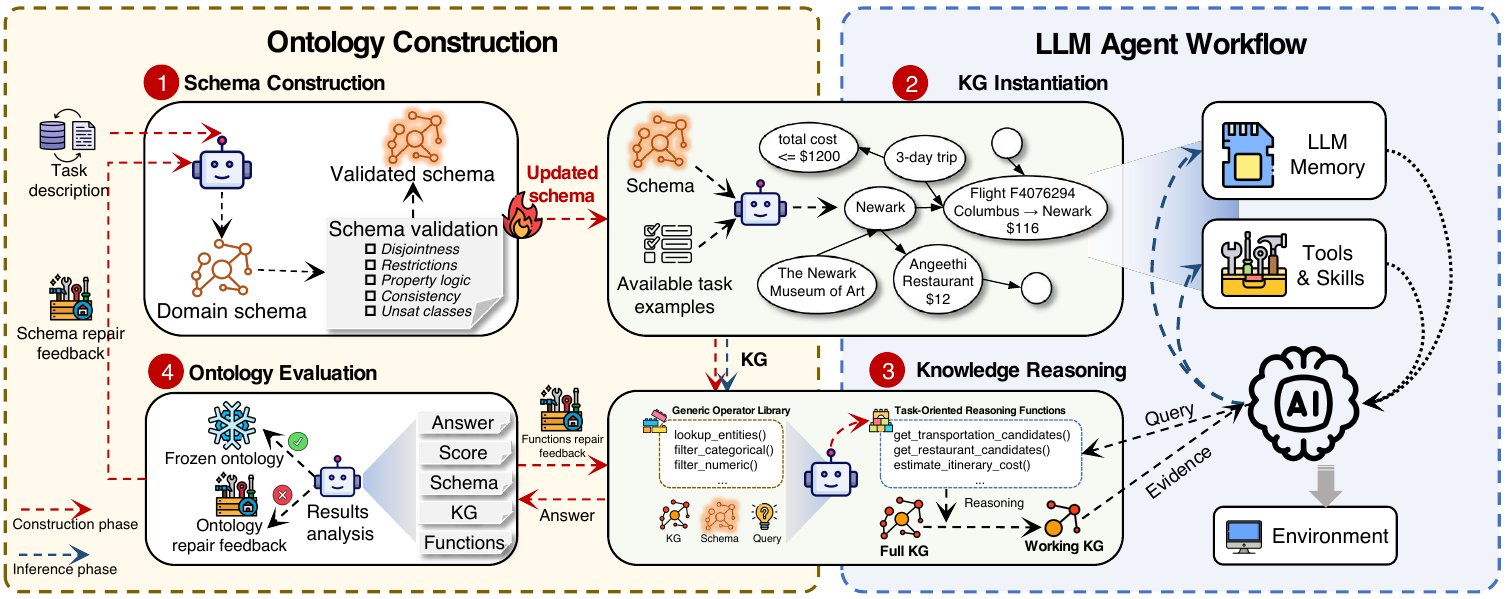}
\caption{Overview of OaK. The construction stage (left) runs a refinement loop over the ontology kernel. The inference stage (right) freezes the kernel and lets a ReAct agent solve unseen queries by calling functions as tools.}
\label{fig:framework}
\end{figure*}

\subsection{Overview}

OaK builds a dynamic task-oriented \emph{ontology kernel} and uses it to mediate between an LLM agent and domain tasks.
The kernel has two main components, namely the schema $\mathcal{S}$ and the function set $\mathcal{F}$.
The schema $\mathcal{S}$ bounds what can be expressed and the functions  $\mathcal{F}$ bound what can be computed.
Once frozen, the kernel is the only channel through which the agent reaches the data. It cannot name a concept or invoke a computation the kernel does not declare.

OaK runs in two stages.
The construction stage runs on training data, where it builds the kernel and refines it in a loop.
The inference stage freezes the kernel and applies it to unseen queries.
Figure~\ref{fig:framework} shows the full loop.

\subsection{Ontology Construction Loop}

The construction stage runs a four-step loop over the kernel.
At the start of each round $t$, OaK draws a fresh random sample $D_t\subseteq D^{\mathrm{tr}}$ from the training set and uses it to drive that round.
Each instance in $D_t$ is a complete example $(q,C_q)$ that pairs a query $q$ with the reference corpus $C_q$ needed to answer it. We write $Q_t=\{q\}$ for its queries and $C_t=\{C_q\}$ for the accompanying corpora.
Resampling per round exposes the loop to varied data and keeps the schema and functions from overfitting to a fixed subset.
We denote by $\mathcal{S}_t$ and $\mathcal{F}_t$ the schema and functions produced in round $t$.

\paragraph{Step 1: Schema construction.}
This step proceeds in three phases: requirement analysis, schema drafting, and formal verification.

\emph{Requirement analysis.}
An LLM reads the task description $T$ together with the round sample $D_t$ and produces a requirement specification:
\[
R=\mathrm{Analyze}(T,D_t).
\]
$R$ records the task scope, the key entities and relations, and the task constraints. It gives the schema a precise target rather than a free form brief.

\emph{Schema drafting.}
The model then drafts the schema from $R$ together with the previous round's schema feedback:
\[
\mathcal{S}_t=\mathrm{Draft}(R,\,\psi^{\mathcal{S}}_{t-1}),
\]
where $\psi^{\mathcal{S}}_{t-1}$ is the schema level repair feedback from the previous round (empty in the first round).
It firstly enumerates entity types with their properties, and then declares the typed relations among them.
Each entity type is assigned an identity field that serves as its primary key, and each relation names its source and target entity types.

\emph{Formal verification.}
A drafted schema can look semantically reasonable yet be logically flawed inside, and such a flaw would propagate into the graph. We therefore verify the draft before it is used.
Our schema is lightweight---a YAML or JSON document of named entity types, properties, and typed relations that the agent reads directly, not the full axiomatic machinery of a formal ontology.
This makes it easy for the model to consume, but a logical reasoner cannot check it as is.
We therefore encode it as an equivalent OWL ontology and run HermiT~\citep{hermit2014}:
\[
\mathrm{HermiT}\!\left(\mathrm{OWL}(\mathcal{S}_t)\right)\rightarrow\{\mbox{consistent},\ \mbox{unsatisfiable classes}\}.
\]
HermiT checks five kinds of logical validity---disjointness, restriction, property-level, and global consistency, plus unsatisfiable classes. We detail them in Appendix~\ref{app:consistency}.
A schema that fails any check is sent back to drafting with the reasoner's counterexamples, and the loop retries.

\paragraph{Step 2: Knowledge Graph instantiation.}
Under the current schema, OaK instantiates a knowledge graph from the reference corpora of the round sample.
To improve extraction quality and avoid exceeding the context limit of the language-model extractor, graph construction uses a chunk--map--merge pipeline.
The corpus $C_t$ is first partitioned into token-bounded chunks $\{c_1,\dots,c_n\}$ that fit the extractor's input budget.
An LLM extractor $\Phi_{\mathcal{S}_t}$ maps each chunk $c_i$ to a set of typed entity and relation candidates according to $\mathcal{S}_t$, and a merge operator $\bigsqcup$ reconciles these sets into one graph:
\[
\mathcal{G}_t=\mathrm{Build}(\mathcal{S}_t,C_t)=\bigsqcup_{i=1}^{n}\Phi_{\mathcal{S}_t}(c_i),
\]

Because $\Phi_{\mathcal{S}_t}$ runs on each chunk independently, one real-world entity may surface as several candidates across chunks.
The merge operator $\bigsqcup$ resolves these duplicates through the schema-declared primary key.
Each candidate entity $e$ is assigned a \emph{key signature}
\[
\kappa(e)=\big(\tau(e),\,\pi_{\mathrm{pk}}(e)\big),
\]
where $\tau(e)$ is its entity type and $\pi_{\mathrm{pk}}(e)$ is its primary-key value.
Two candidates denote the same real-world object exactly when their key signatures agree:
\[
e_i \sim e_j \iff \kappa(e_i)=\kappa(e_j).
\]
Since $\sim$ is an equivalence relation, $\bigsqcup$ collapses each equivalence class $[e]_\sim$ into one canonical node with a single deterministic identifier.
Relations are then re-attached to these canonical endpoints. This process also collapses duplicate relations produced across chunks.

\paragraph{Step 3: Knowledge reasoning.}
This step composes the generic operators into a catalog of domain-adapted functions. 
A function compiles a task's recurring reasoning steps into a single typed call. 
This frees the agent from multi-step planning. 
With less reasoning done inside the model, there is less room for hallucination.
An LLM-based composer assembles the catalog $\mathcal{F}_t$:
\[
\mathcal{F}_t=\mathrm{Compose}\big(\mathcal{O},\,\mathcal{S}_t,\,\mathcal{G}_t,\,Q_t,\,\psi^{\mathcal{F}}_{t-1}\big),
\]
where $\psi^{\mathcal{F}}_{t-1}$ is the function-level repair feedback from the previous round (empty in the first round).
By reading how queries in $Q_t$ resolve over $\mathcal{G}_t$, the composer identifies the recurring reasoning patterns the functions must cover.

Each pattern is realized as one function $f\in\mathcal{F}_t$, with typed inputs and outputs and a realization $r_f$ over $\mathcal{O}$.
The generic library $\mathcal{O}$ provides basic graph and data operations, including entity lookup, relation traversal, property projection, constraint filtering, and aggregation. Appendix~\ref{app:operators} gives the complete operator signatures and descriptions.
A function is grounded in $\mathcal{O}$ in one of three ways: it may \emph{compose} several operators into a pipeline, \emph{specialize} an operator by fixing or reinterpreting its parameters, or \emph{adapt} an operator with lightweight pre- or post-processing.
Each function is tested on $\mathcal{G}_t$ to confirm it is executable, and the validated catalog is exposed to the agent as callable tools.

With the kernel in place, a ReAct agent runs on the queries:
\[
A_t=\mathrm{Agent}(Q_t,\mathcal{S}_t,\mathcal{G}_t,\mathcal{F}_t).
\]
For each query in $Q_t$, the agent invokes the functions $\mathcal{F}_t$ as tools, binds their typed arguments according to the schema $\mathcal{S}_t$, and executes them over the graph $\mathcal{G}_t$ to retrieve the evidence needed to answer the query.
The trajectory $A_t$ collects the resulting operator traces, function outputs and final answers, which the next step reviews.

\paragraph{Step 4: Ontology evaluation.}
The final answers in $A_t$ are first scored according to the dataset's evaluation protocol:
\[
\mathbf{s}_t=\mathrm{Eval}(A_t).
\]
A judge model then reviews the whole kernel together with the trajectory and its task scores~\citep{llmjudge2023}:
\[
\psi_t=\mathrm{Judge}(\mathcal{S}_t,\mathcal{G}_t,\mathcal{F}_t,A_t,\mathbf{s}_t).
\]
$\psi_t$ is a set of repair suggestions over the round's artifacts.
Let $\mathcal{U}_t = E_t \cup R_t \cup \mathcal{F}_t$ collect the entity types $E_t$, relation types $R_t$, and functions $\mathcal{F}_t$ of round $t$, and let $\mathrm{Act}=\{\textsf{add},\textsf{delete},\textsf{modify}\}$.
Each suggestion is a tuple
\[
\sigma = (u,\, a,\, \delta,\, \rho),\qquad u\in\mathcal{U}_t,\ a\in\mathrm{Act},
\]
where $u$ is the target artifact, $a$ the action, $\delta$ the proposed definition or patch, and $\rho$ the diagnosed reason.
The report is
\[
\psi_t = \{\sigma_1,\dots,\sigma_k\},
\]
partitioned by target into $\psi_t = \psi_t^{E}\cup \psi_t^{R}\cup \psi_t^{\mathcal{F}}$, with the schema level part $\psi_t^{\mathcal{S}}=\psi_t^{E}\cup \psi_t^{R}$ feeding Step 1 and the function level part $\psi_t^{\mathcal{F}}$ feeding Step 3 of round $t{+}1$.

OaK then applies the repair and starts the next round.
The loop stops when the judge finds no blocking fault or the iteration budget runs out.

\subsection{Ontology-driven LLM Inference}

Inference runs on the unseen test set, with no further edits to the kernel.
After the loop, OaK freezes the schema $\mathcal{S}^*$ and the functions $\mathcal{F}^*$ from the final round.
For a test query $q$, it builds an inference graph:
\[
\mathcal{G}_q=\mathrm{Build}(\mathcal{S}^*,C_q),
\]
where $C_q$ is the corpus that accompanies $q$ in test set.

A ReAct agent then solves $q$ with the frozen kernel:
\[
a=\mathrm{Agent}(q,\mathcal{S}^*,\mathcal{G}_q,\mathcal{F}^*).
\]
As in construction, the agent invokes $\mathcal{F}^*$ as tools, binds their typed arguments under $\mathcal{S}^*$, and executes them over $\mathcal{G}_q$.
Unlike free-form agent reasoning, inference here is routed through a task-refined kernel: $\mathcal{S}^*$ closes off concepts and relations the agent may invoke, and $\mathcal{F}^*$ restricts its computations to typed, executable compositions over schema-constrained graph instances.
The agent therefore operates inside a verified semantic space: answers are meant to trace to grounded evidence rather than to model-internal inference.
This can reduce reasoning errors and unsupported inferences in open-ended generation.

\section{Experiments}

\begin{table}[t]
\centering
\small
\setlength{\tabcolsep}{5pt}
\renewcommand{\arraystretch}{1.12}
\resizebox{\columnwidth}{!}{
\begin{tabular}{llccccc}
\toprule
\multicolumn{1}{c}{LLM} & \multicolumn{1}{c}{Method} & \multicolumn{2}{c}{CS} & \multicolumn{2}{c}{HC} & Final \\
\cmidrule(lr){3-4}\cmidrule(lr){5-6}
& & Micro & Macro & Micro & Macro &  \\
\midrule
\multirow{6}{*}{\rotatebox{90}{\textbf{DeepSeek-v4-flash}}} & ReAct       & 81.60 & 19.10 & 44.29 & 36.40 & 15.30 \\
& AFlow       & 79.31 & 31.70 & 43.60 & 42.70 & 29.10 \\
& MemP        & 82.78 & \underline{55.50} & \textbf{63.91} & \underline{59.50} & \underline{51.50} \\
& ReCode      & \textbf{86.60} & 50.90 & 48.67 & 37.60 & 48.20 \\
& AgentSquare & 79.10 & 30.40 & 49.40 & 47.80 & 27.70 \\
& OaK         & \underline{86.11} & \textbf{58.60} & \underline{59.31} & \textbf{61.57} & \textbf{55.90} \\
\midrule
\multirow{6}{*}{\rotatebox{90}{\textbf{GPT-4o-mini}}} & ReAct       & 79.30 & 14.20 & 29.20 & 17.80 & 4.50 \\
& AFlow       & 78.23 & 12.60 & 31.09 & 17.70 & 3.00 \\
& MemP        & 68.43 & 17.40 & 27.29 & 19.50 & \underline{16.30} \\
& ReCode      & \textbf{88.59} & \underline{22.00} & \underline{54.50} & \underline{26.50} & 15.00 \\
& AgentSquare & 72.06 & 13.50 & 21.11 & 12.10 & 5.40 \\
& OaK         & \underline{82.40} & \textbf{30.80} & \textbf{68.00} & \textbf{48.30} & \textbf{19.70} \\
\bottomrule
\end{tabular}
}
\caption{TravelPlanner results on DeepSeek-v4-flash and gpt-4o-mini.
All entries are percentages (\%).
CS = Commonsense Constraint Pass Rate; HC = Hard Constraint Pass Rate; Final = Final Pass Rate.
The best value in each column is in \textbf{bold} and the second best is \underline{underlined}.}
\label{tab:travelplanner}
\end{table}
\subsection{Tasks}

We choose these three benchmarks. They cover complementary agent settings, including multi-step planning, CRM workflow execution, and compositional tool use over heterogeneous corpora.
This diversity lets us evaluate OaK's applicability across multiple task settings.
For every dataset, we follow the official evaluation protocol and metrics.

\begin{itemize}
    \item \textbf{TravelPlanner}~\citep{travelplanner2024} evaluates multi-day plans that jointly arrange transportation, meals, attractions, and accommodation.
    Commonsense (CS) and hard-constraint (HC) measure overall feasibility and explicit requirement compliance, respectively. 
    Within each family, micro is the fraction of individual constraints satisfied, macro is the fraction of plans meeting all applicable constraints, and final requires both families to be fully satisfied.
    \item \textbf{CRMArenaPro}~\citep{crmarenaPro2025} evaluates Customer Relationship Management (CRM) tasks in synthetic B2B and B2C organizations. These represent company-oriented and individual-customer processes, respectively.
    Workflow, Policy, and Database use exact match, whereas Text uses exact match for discrete answers or token-level F1 for free-form answers.
    \item \textbf{ToolQA}~\citep{toolqa2023} evaluates compositional tool use over heterogeneous external corpora using normalized exact match.
\end{itemize}

\subsection{Implementation Details}

We run every method on two LLM backbones, \texttt{DeepSeek-v4-flash} and \texttt{gpt-4o-mini}.
These backbones are used throughout Step 1's requirement analysis and schema drafting, Step 2's knowledge graph instantiation, Step 3's function composition and knowledge reasoning. 
The ontology evaluator is held fixed across all settings and uses \texttt{claude-sonnet-4.6}.
Each construction round draws a fresh sample covering 20\% of the training split, 
and the loop runs for at most 5 iterations.
The ReAct agent is capped at 20 steps per query.

\subsection{Main Results}

\paragraph{TravelPlanner.}
In Table~\ref{tab:travelplanner}, OaK achieves the highest final pass rate under both backbones and leads all macro-level metrics on the test sets.
On the DeepSeek-v4-flash backbone, ReCode is strongest on commonsense micro scores and MemP remains competitive on hard-constraint micro scores. This indicates that satisfying individual constraints does not necessarily produce a jointly valid plan.
TravelPlanner requires decisions about dates, cities, transportation, accommodation, dining, and budgets to remain compatible across multiple days. This makes cross-component coordination essential.
OaK explicitly represents these entities and constraints in its schema. Its knowledge graph connects them to the available options.
Its schema-adapted functions further integrate candidate retrieval, budget calculation, and constraint checking into a coherent planning procedure.
This combination helps OaK preserve dependencies across the complete itinerary, explaining its stronger macro and final performance rather than merely improving isolated constraint satisfaction.

\begin{table*}[t]
\centering
\small
\setlength{\tabcolsep}{5pt}
\renewcommand{\arraystretch}{1.12}
\resizebox{\textwidth}{!}{
\begin{tabular}{llccccc@{\hspace{0.45cm}}ccccc}
\toprule
\multicolumn{1}{c}{LLM} & \multicolumn{1}{c}{Method} & \multicolumn{5}{c}{B2B} & \multicolumn{5}{c}{B2C} \\
\cmidrule(lr){3-7}\cmidrule(lr){8-12}
& & Workflow & Policy & Text & Database & Avg. & Workflow & Policy & Text & Database & Avg. \\
\midrule
\multirow{6}{*}{\rotatebox{90}{\textbf{DeepSeek-v4-flash}}} & ReAct & 92.50 & 47.50 & 31.57 & 71.56 & 60.78 & 88.75 & 45.62 & \underline{43.47} & 65.62 & 60.87 \\
& AFlow & \underline{96.25} & 46.25 & 33.87 & 48.44 & 56.20 & 93.75 & 47.50 & 41.84 & 33.75 & 54.21 \\
& MemP & \underline{96.25} & \underline{55.62} & \underline{35.74} & \underline{78.12} & \underline{66.44} & \underline{95.00} & \underline{61.25} & \textbf{43.77} & \underline{69.06} & \underline{67.27} \\
& ReCode & 26.25 & 39.38 & 10.13 & 24.06 & 24.96 & 27.50 & 34.38 & 11.71 & 24.06 & 24.41 \\
& AgentSquare & 95.00 & 46.88 & 31.10 & 50.94 & 55.98 & 88.75 & 56.25 & 36.37 & 44.06 & 56.36 \\
& OaK & \textbf{97.50} & \textbf{81.88} & \textbf{39.46} & \textbf{94.69} & \textbf{78.38} & \textbf{100.00} & \textbf{71.88} & 39.87 & \textbf{89.06} & \textbf{75.20} \\
\midrule
\multirow{6}{*}{\rotatebox{90}{\textbf{GPT-4o-mini}}} & ReAct & 27.50 & 27.50 & 7.73 & 17.81 & 20.14 & 15.00 & 33.13 & 13.80 & 23.12 & 21.28 \\
& AFlow & 66.25 & 46.25 & 5.61 & 25.62 & 35.93 & 53.75 & 41.25 & 8.80 & 24.38 & 32.04 \\
& MemP & \underline{71.25} & 45.63 & \underline{30.73} & \underline{49.38} & \underline{49.24} & \underline{82.50} & \underline{44.38} & \underline{31.11} & \underline{41.56} & \underline{49.89} \\
& ReCode & 21.75 & 38.12 & 12.00 & 16.88 & 22.18 & 23.25 & 33.75 & 11.04 & 16.56 & 21.15 \\
& AgentSquare & 16.25 & \underline{48.13} & 5.01 & 31.56 & 25.24 & 17.50 & 40.62 & 7.88 & 28.75 & 23.69 \\
& OaK & \textbf{83.75} & \textbf{55.00} & \textbf{32.27} & \textbf{84.69} & \textbf{63.93} & \textbf{87.50} & \textbf{55.63} & \textbf{40.80} & \textbf{86.56} & \textbf{67.62} \\
\bottomrule
\end{tabular}
}
\caption{CRMArenaPro results on DeepSeek-v4-flash and gpt-4o-mini.
All entries are percentages (\%); Workflow, Policy, Text, and Database are task-category scores.
Avg. is the equal-weight mean of Workflow, Policy, Text, and Database.
The best value in each column is in \textbf{bold} and the second best is \underline{underlined}.}
\label{tab:crmarenapro}
\end{table*}

\paragraph{CRMArenaPro.}
On CRMArenaPro, Table~\ref{tab:crmarenapro} reports the highest Avg. for OaK in both B2B and B2C organizations under both backbones, and leads in nearly all category-level comparisons.
Its advantages are particularly clear on Policy and Database tasks. They require agents to combine business-rule compliance with accurate access to records.
Workflow-oriented baselines do well on some Workflow tasks, and MemP shows the value of reusable procedures. 
However, their performance drops when the task requires selecting the correct records, fields, and relations.
OaK addresses this limitation by explicitly representing CRM record types, fields, and their relations.
It also grounds reusable functions for cross-table queries, rule checks, and numerical calculations in this representation.
Its larger advantage with \texttt{gpt-4o-mini}, especially on Database tasks, further suggests that schema-grounded execution reduces the amount of multi-step data reasoning that must be performed by the backbone model alone.

\begin{table}[t]
\centering
\small
\setlength{\tabcolsep}{5pt}
\renewcommand{\arraystretch}{1.12}
\resizebox{\columnwidth}{!}{
\begin{tabular}{llcccccc}
\toprule
LLM & Method & Flight & Coffee & Airbnb & DBLP & Yelp & Avg. \\
\midrule
\multirow{6}{*}{\rotatebox{90}{\textbf{DeepSeek-v4-flash}}} & AFlow & \underline{57.78} & 49.28 & 53.89 & 35.56 & 70.00 & 53.18 \\
& AgentSquare & 46.67 & 41.06 & 50.56 & 28.89 & 63.89 & 46.06 \\
& ReAct & 37.78 & 48.79 & 48.33 & \underline{43.18} & 82.20 & 51.96 \\
& ReCode & 28.95 & 18.40 & 46.15 & 12.50 & 57.14 & 32.21 \\
& MemP & 55.56 & 48.79 & \underline{54.44} & 35.00 & \underline{82.22} & \underline{55.02} \\
& OaK & \textbf{66.67} & \textbf{64.25} & \textbf{58.33} & \textbf{44.94} & \textbf{88.76} & \textbf{64.58} \\
\midrule
\multirow{6}{*}{\rotatebox{90}{\textbf{GPT-4o-mini}}} & AFlow & 30.00 & 38.16 & 52.78 & \underline{30.00} & 57.78 & 41.64 \\
& AgentSquare & 10.00 & 40.58 & 42.78 & 16.11 & 35.56 & 29.34 \\
& ReAct & \textbf{37.38} & \underline{63.54} & \textbf{72.28} & 18.92 & \underline{58.18} & \underline{50.45} \\
& ReCode & 17.76 & 62.39 & 52.86 & 12.90 & 42.74 & 38.45 \\
& MemP & 12.22 & 30.43 & 46.67 & 17.22 & 33.33 & 28.05 \\
& OaK & \underline{31.10} & \textbf{82.60} & \underline{54.40} & \textbf{43.30} & \textbf{81.70} & \textbf{59.32} \\
\bottomrule
\end{tabular}
}
\caption{ToolQA results on the five subsets, using DeepSeek-v4-flash and gpt-4o-mini as backbones.
All entries are percentages (\%) and report exact-match accuracy.
Avg. is the weighted average across subsets. 
The best value in each column is in \textbf{bold} and the second best is \underline{underlined}.}
\label{tab:toolqa}
\end{table}

\paragraph{ToolQA.}
For ToolQA, Table~\ref{tab:toolqa} reports the highest weighted-average performance for OaK under both backbones, with first-place results on most subset--backbone combinations.
ToolQA requires agents to compose multiple operations over heterogeneous corpora, including field lookup, record filtering, relation traversal, and aggregation.
Success therefore depends on coordinating the order of operations with the semantics of domain-specific fields and relations, rather than selecting tools independently.
OaK addresses this requirement by mapping each question to typed and reusable procedures whose arguments and operations are constrained by an explicit domain schema and grounded knowledge graph.
Compared with methods that optimize global workflows or procedural reuse without explicitly representing data relations, OaK more reliably connects each operation with the appropriate records and relations.

Additional robustness results over multiple random seeds are provided in Appendix~\ref{app:robustness}.

\begin{figure}[t]
\centering
\includegraphics[width=\columnwidth]{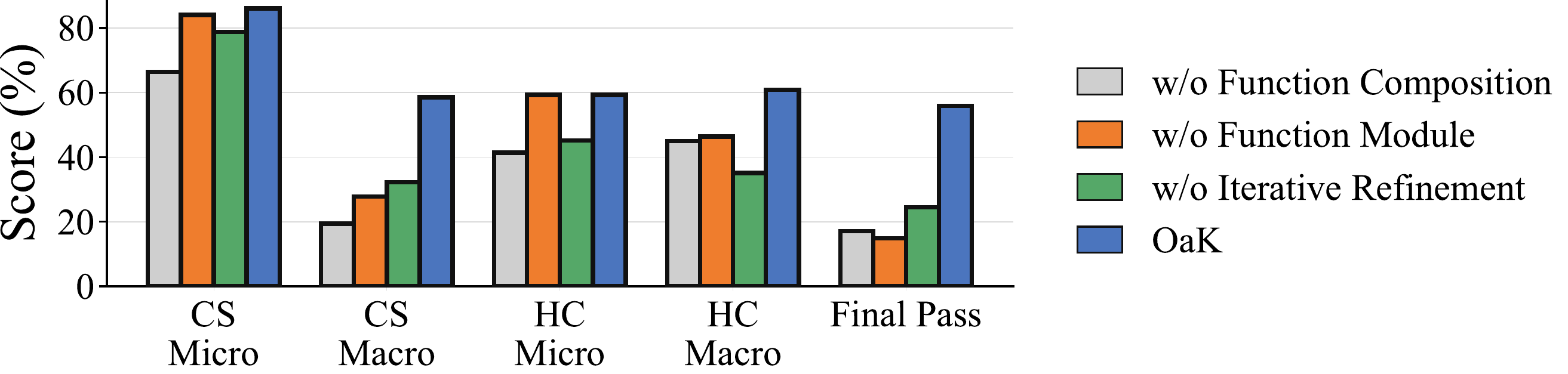}
\caption{Ablation results on TravelPlanner.}
\label{fig:travelplanner-ablation}
\end{figure}

\subsection{Ablation Study}

We compare the full OaK with three variants using DeepSeek-v4-flash.
We provide the ablations using gpt-4o-mini in Appendix~\ref{app:ablation}.
\emph{w/o Function Composition} removes the composition step and exposes the generic operators $\mathcal{O}$ directly to the agent.
\emph{w/o Function Module} removes $\mathcal{F}$ entirely and lets the LLM reason over the full graph, retaining only minimal retrieval operations when the graph cannot be placed directly in context.
\emph{w/o Iterative Refinement} executes only the first round of the construction loop, without subsequent updates.

\paragraph{TravelPlanner.}
In Figure~\ref{fig:travelplanner-ablation}, removing any component substantially reduces the final pass rate. The largest degradation is caused by removing the function module or function composition.
Although direct graph access can preserve individual hard-constraint decisions, it does not provide reusable procedures for coordinating transportation, accommodation, dining, and budget constraints across a complete plan.
Without function composition, the agent must reconstruct these dependent operations for each query. This further weakens global plan consistency.
The one-round variant also performs substantially worse. This shows that iterative refinement is necessary for repairing missing constraints and incomplete reasoning procedures.

\begin{figure}[t]
\centering
\includegraphics[width=0.8\columnwidth]{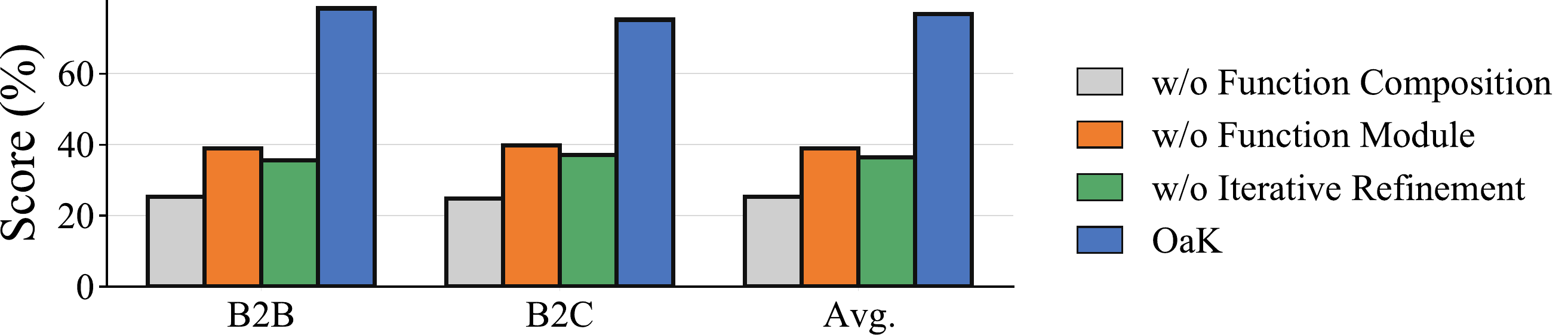}
\caption{Ablation results on CRMArenaPro.}
\label{fig:crmarenapro-ablation}
\end{figure}

\paragraph{CRMArenaPro.}
In Figure~\ref{fig:crmarenapro-ablation}, the full OaK model outperforms all variants on both B2B and B2C settings.
Removing function composition causes the largest degradation because the agent must reconstruct cross-table queries, business-rule checks and numerical calculations from generic operators.
Removing the function module also reduces performance by eliminating reusable procedures for recurring CRM operations, even when graph access is available.

\begin{figure}[h]
\centering
\includegraphics[width=\columnwidth]{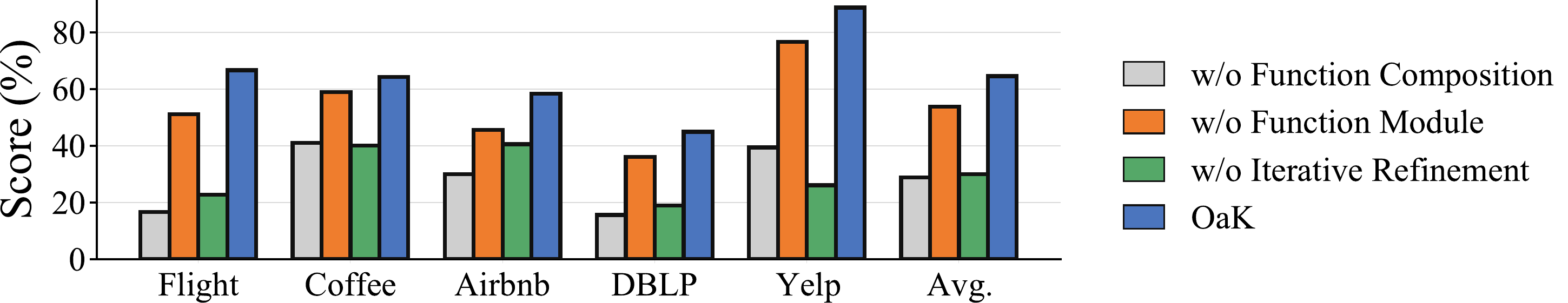}
\caption{Ablation results on ToolQA.}
\label{fig:toolqa-ablation}
\end{figure}

\paragraph{ToolQA.}
Across all subsets in Figure~\ref{fig:toolqa-ablation}, OaK outperforms all variants.
The function-module ablation is the strongest variant but still remains clearly below the full model. This indicates that graph access alone cannot replace schema-adapted procedures.
Removing function composition substantially harms performance because the agent must reconstruct multi-step sequences of lookup, filtering, relation traversal, projection, and aggregation for each question.

\subsection{Construction Loop Analysis}

\begin{figure}[t]
\centering
\includegraphics[width=\columnwidth]{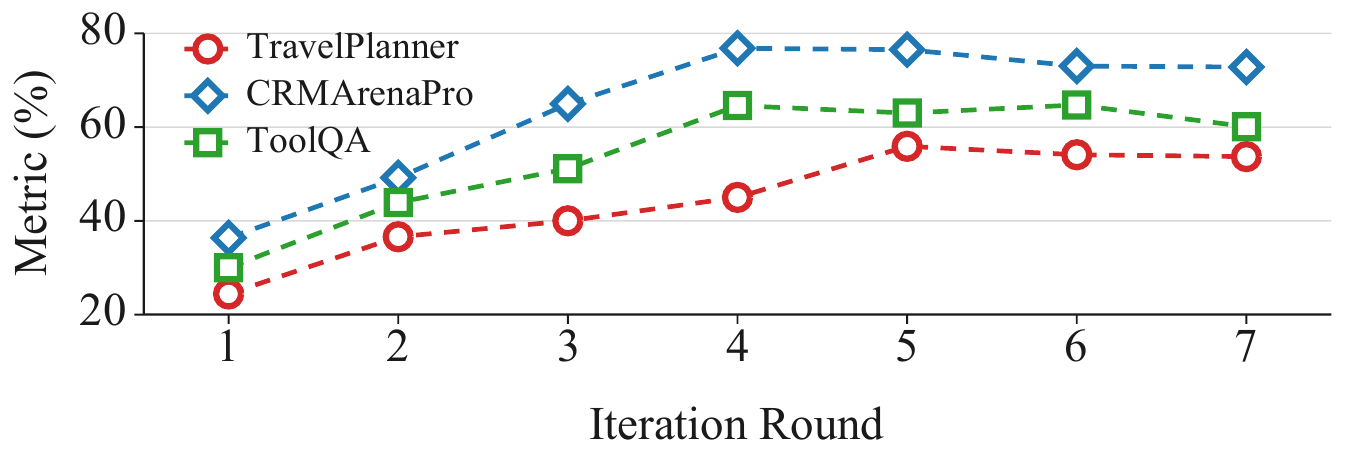}
\caption{Loop progress on TravelPlanner, CRMArenaPro, and ToolQA using DeepSeek-v4-flash.}
\label{fig:iteration-progress}
\end{figure}

We record each benchmark's primary aggregate metric after every construction round under a fixed inference protocol.
Across all three benchmarks, Figure~\ref{fig:iteration-progress} shows rapid improvement during the early rounds. This indicates that judge-guided updates quickly fix missing constraints and graph-mapping defects while closing function-level gaps.
The curves largely plateau after rounds 4--5, suggesting that the five-round budget captures most of the benefit of iterative refinement.
The small fluctuations in later rounds likely result from fine-tuning of the schema and functions, together with the fresh training sample used in each round.

\subsection{Case Study}

We illustrate how judge feedback coordinates schema and function level repair in TravelPlanner.

\paragraph{Schema repair.}
\begin{figure}[t]
\centering
\begingroup
\small
\setlength{\fboxsep}{0pt}
\fcolorbox{black}{codebg}{%
\begin{tabular}{@{}l@{}}
\ttfamily\strut\hspace{0.6em}\makebox[1.2em][l]{}\hspace{0.6em}name: Hotel\\
\ttfamily\strut\hspace{0.6em}\makebox[1.2em][l]{}\hspace{0.6em}primary\_key: name\\
\ttfamily\strut\hspace{0.6em}\makebox[1.2em][l]{}\hspace{0.6em}properties:\\
\ttfamily\strut\hspace{0.6em}\makebox[1.2em][l]{}\hspace{0.6em}\hspace{1.2em}-\hspace{0.6em}name: name\\
\ttfamily\strut\hspace{0.6em}\makebox[1.2em][l]{}\hspace{0.6em}\hspace{1.2em}-\hspace{0.6em}name: city\\
\ttfamily\strut\hspace{0.6em}\makebox[1.2em][l]{}\hspace{0.6em}\hspace{1.2em}-\hspace{0.6em}name: room\_type\\
\ttfamily\strut\hspace{0.6em}\makebox[1.2em][l]{}\hspace{0.6em}\hspace{1.2em}-\hspace{0.6em}name: price\_per\_night\\
\rowcolor{diffaddbg}[0pt][0pt]\color{black}\ttfamily\strut\hspace{0.6em}\makebox[1.2em][l]{+}\hspace{0.6em}\hspace{1.2em}-\hspace{0.6em}name: minimum\_nights\\
\rowcolor{diffaddbg}[0pt][0pt]\color{black}\ttfamily\strut\hspace{0.6em}\makebox[1.2em][l]{+}\hspace{0.6em}\hspace{1.2em}-\hspace{0.6em}name: maximum\_occupancy\\
\end{tabular}%
}
\endgroup
\caption{Schema repair for the \texttt{Hotel} entity type.}
\label{fig:schemarepair}
\end{figure}
In an early construction round, the judge found that the \texttt{Hotel} entity exposed visible attributes such as price and room type but omitted the \texttt{minimum\_nights} and \texttt{maximum\_occupancy} constraints.
Without these fields, the agent could select a hotel that appeared inexpensive and suitable but was invalid because it required a longer stay or could not accommodate the entire party.
The judge therefore issued an \emph{add} suggestion for the \texttt{Hotel} entity, adding both constraint fields as shown in Figure~\ref{fig:schemarepair}.
The updated schema made these requirements explicit and allowed the agent to incorporate them during planning.

\paragraph{Function repair.}

The schema update alone was insufficient as the \texttt{get\_accommodation\_candidates} function did not yet use the newly exposed constraints.
The ontology evaluator traced remaining failed plans and issued a \emph{modify} suggestion that added filters for number of people and minimum stay, as shown in Figure~\ref{fig:methodrepair}.
After the repair, a hotel was retained only when its maximum occupancy covered the party size and its minimum-night requirement fit the planned stay.

\begin{figure}[t]
\centering
\begingroup
\small
\setlength{\fboxsep}{0pt}
\fcolorbox{black}{codebg}{%
\begin{tabular}{@{}l@{}}
\ttfamily\strut\hspace{0.6em}\makebox[1.2em][l]{}\hspace{0.6em}def get\_accommodation\_candidates(\\
\ttfamily\strut\hspace{0.6em}\makebox[1.2em][l]{}\hspace{0.6em}\hspace{1.2em}hotels, budget, people\_count,\\
\ttfamily\strut\hspace{0.6em}\makebox[1.2em][l]{}\hspace{0.6em}\hspace{1.2em}stay\_nights, room\_type):\\
\ttfamily\strut\hspace{0.6em}\makebox[1.2em][l]{}\hspace{0.6em}\hspace{1.2em}hotels = filter\_numeric(\\
\ttfamily\strut\hspace{0.6em}\makebox[1.2em][l]{}\hspace{0.6em}\hspace{2.4em}hotels, "price\_per\_night",\\
\ttfamily\strut\hspace{0.6em}\makebox[1.2em][l]{}\hspace{0.6em}\hspace{2.4em}"\textless=", budget)\\
\ttfamily\strut\hspace{0.6em}\makebox[1.2em][l]{}\hspace{0.6em}\hspace{1.2em}hotels = filter\_categorical(\\
\ttfamily\strut\hspace{0.6em}\makebox[1.2em][l]{}\hspace{0.6em}\hspace{2.4em}hotels, "room\_type", room\_type)\\
\rowcolor{diffaddbg}[0pt][0pt]\color{black}\ttfamily\strut\hspace{0.6em}\makebox[1.2em][l]{+}\hspace{0.6em}\hspace{1.2em}hotels = filter\_numeric(\\
\rowcolor{diffaddbg}[0pt][0pt]\color{black}\ttfamily\strut\hspace{0.6em}\makebox[1.2em][l]{}\hspace{0.6em}\hspace{2.4em}hotels, "maximum\_occupancy",\\
\rowcolor{diffaddbg}[0pt][0pt]\color{black}\ttfamily\strut\hspace{0.6em}\makebox[1.2em][l]{}\hspace{0.6em}\hspace{2.4em}"\textgreater=", people\_count)\\
\rowcolor{diffaddbg}[0pt][0pt]\color{black}\ttfamily\strut\hspace{0.6em}\makebox[1.2em][l]{+}\hspace{0.6em}\hspace{1.2em}hotels = filter\_numeric(\\
\rowcolor{diffaddbg}[0pt][0pt]\color{black}\ttfamily\strut\hspace{0.6em}\makebox[1.2em][l]{}\hspace{0.6em}\hspace{2.4em}hotels, "minimum\_nights",\\
\rowcolor{diffaddbg}[0pt][0pt]\color{black}\ttfamily\strut\hspace{0.6em}\makebox[1.2em][l]{}\hspace{0.6em}\hspace{2.4em}"\textless=", stay\_nights)\\
\ttfamily\strut\hspace{0.6em}\makebox[1.2em][l]{}\hspace{0.6em}return rank\_and\_select(hotels)\\
\end{tabular}%
}
\endgroup
\caption{Function repair for \texttt{get\_accommodation\_\protect\\ candidates}.}
\label{fig:methodrepair}
\end{figure}

\subsection{Cost Analysis}

\begin{figure}[H]
\centering
\includegraphics[width=\columnwidth]{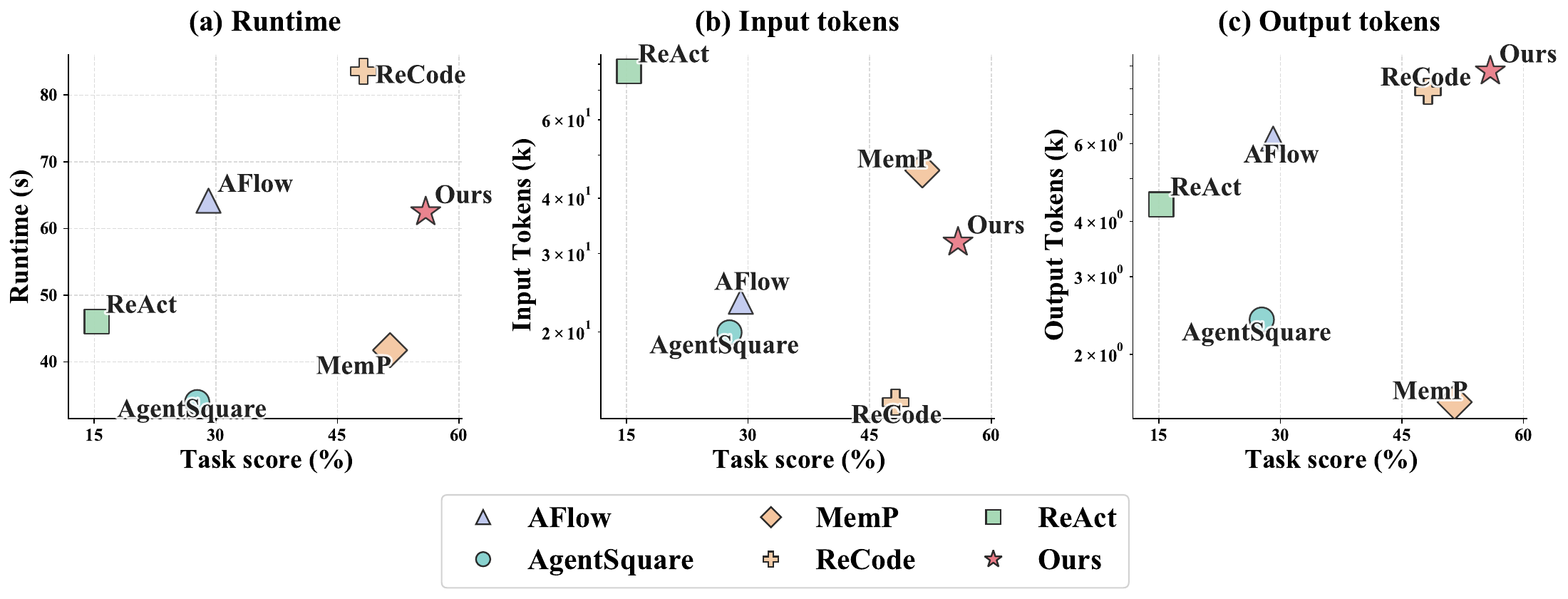}
\caption{Resource cost versus final pass rate on TravelPlanner using DeepSeek-v4-flash.}
\label{fig:cost-analysis}
\end{figure}

We compare OaK with the baselines on TravelPlanner using runtime and token counts as resource proxies and final pass rate as the task-quality measure.
In Figure~\ref{fig:cost-analysis}, OaK achieves the highest final pass rate while using fewer input tokens than ReAct and MemP.
This input reduction suggests that schema-adapted functions reduce the need to reconstruct graph operations from a large context.
The benefit comes with higher runtime and output-token usage than several baselines, mainly because OaK instantiates a query-specific graph before planning.
When a graph can be reused across related queries, this construction cost can be amortized.
Overall, OaK trades additional execution and output cost for stronger task performance while keeping input-context cost moderate.

\section{Conclusion and Future Work}

We presented OaK, an ontology-as-a-kernel framework that turns domain semantics into an executable interface for LLM agents.
OaK constructs a task-oriented schema and instantiates a schema-guided knowledge graph. It then composes typed reasoning functions from generic operators over this graph. Official task scores and execution traces guide iterative repairs.
Across TravelPlanner, CRMArenaPro and ToolQA, OaK achieves the best aggregate performance with two LLM backbones. The ablations confirm the importance of the function module, function composition, and iterative refinement. 
These results suggest that making domain semantics and reasoning procedures explicit can improve both effectiveness and inspectability in LLM agents.

OaK still pays a graph-instantiation cost.
Its quality also depends on the LLM-based extractor and judge, and requires a reliable task evaluator.
Future work can study reusable and incrementally updated graphs to amortize construction, stronger verification or human feedback for open-ended tasks.
It is also promising to include richer operator libraries and ontology representations for larger and evolving domains.

\bibliography{aaai2027}

\clearpage
\appendix
\setcounter{secnumdepth}{1}

\section{Schema Consistency Checks}
\label{app:consistency}

This appendix details the five kinds of logical validity that the HermiT reasoner verifies on the OWL ontology $\mathrm{OWL}(\mathcal{S}_t)$ 

\paragraph{Disjointness consistency.}
The reasoner checks that the schema does not force categories that should be mutually exclusive to overlap.
For example, a schema should not imply that \textit{Restaurant} is a subclass of \textit{City}.

\paragraph{Restriction consistency.}
The reasoner checks that existential, universal, and cardinality constraints do not jointly create contradictions.
For example, ``a trip has at least one day'' and ``a trip can have zero days'' should not be accepted together.

\paragraph{Property-level consistency.}
The reasoner checks the logical features declared on properties, including functional, inverse-functional, transitive, symmetric, asymmetric, reflexive, irreflexive, and inverse properties. It also verifies that these features remain compatible when combined with the other axioms of the schema.

\paragraph{Global consistency.}
The reasoner checks that the schema remains logically satisfiable as a whole, rather than only in isolated fragments.

\paragraph{Unsatisfiable classes.}
Finally, the reasoner detects unsatisfiable classes, that is, schema components that can never hold an instance once type definitions, relation definitions, and constraints are combined.

\section{Generic Operator Library}
\label{app:operators}

The generic library $\mathcal{O}$ contains the following nine public operators. Underscore-prefixed helpers are implementation details.
\lstdefinestyle{operatorsignature}{
    language=Python,
    basicstyle=\ttfamily\footnotesize,
    columns=fullflexible,
    numbers=none,
    xleftmargin=0pt,
    breaklines=true,
    breakatwhitespace=true,
    showstringspaces=false,
    aboveskip=0.2em,
    belowskip=0.2em
}

\begin{itemize}
\setlength{\itemsep}{0.6em}

\item \textbf{Runtime-slot extraction.}
\begin{lstlisting}[style=operatorsignature]
def extract_runtime_slots(
    *,
    query: str,
    slot_specs: Sequence[RuntimeSlotSpec],
    env_path: str | Path = (
        "/ontology_research/.env"
    ),
    log_dir: str | Path = (
        "ontology_llm_logs"
    ),
    task_context: str = "",
    max_attempts: int = 3,
    max_tokens: int = 2000,
) -> RuntimeSlotExtractionResult:
\end{lstlisting}
Maps a natural-language query to declared typed slots with an LLM, retaining only explicit or strongly implied constraints and their evidence.

\item \textbf{Entity lookup.}
\begin{lstlisting}[style=operatorsignature]
def lookup_entities(
    graph: str | Path | dict[str, Any] | InstantiatedGraph,
    *,
    entity_types: Sequence[str] | None = None,
    entity_ids: Sequence[str] | None = None,
    primary_key: dict[str, Any] | None = None,
    property_filters: dict[str, Any] | None = None,
    name_query: str | None = None,
    text_query: str | None = None,
    fuzzy: bool = True,
    top_k: int | None = None,
    min_score: float = 0.0,
) -> EntityLookupResult:
\end{lstlisting}
Retrieves entities by type, identifier, primary key, exact property values, or fuzzy name and text matching, with optional ranking and truncation.

\item \textbf{Relation traversal.}
\begin{lstlisting}[style=operatorsignature]
def traverse_relations(
    graph: str | Path | dict[str, Any] | InstantiatedGraph,
    *,
    start_entity_ids: Sequence[str],
    relation_types: Sequence[str] | None = None,
    direction: TraversalDirection = "outgoing",
    max_hops: int = 1,
    include_starting_entities: bool = True,
) -> TraversalResult:
\end{lstlisting}
Expands a bounded neighborhood from seed entities along selected relation types and directions, returning the visited entities and relations.

\item \textbf{Property projection.}
\begin{lstlisting}[style=operatorsignature]
def project_properties(
    entities: Sequence[GraphEntity] | EntityLookupResult | EntityFilterResult | TraversalResult,
    *,
    property_names: Sequence[str],
    missing_value: Any = None,
) -> list[ProjectedRow]:
\end{lstlisting}
Projects selected entity properties into flat rows while preserving entity identifiers, types, and primary keys.

\item \textbf{Categorical filtering.}
\begin{lstlisting}[style=operatorsignature]
def filter_property_categorical(
    entities: Sequence[GraphEntity] | EntityLookupResult | EntityFilterResult | TraversalResult,
    *,
    conditions: Sequence[
    PropertyCategoricalFilterCondition
    ],
) -> EntityFilterResult:
\end{lstlisting}
Keeps entities that satisfy exact categorical inclusion or exclusion conditions. $\mathcal{O}$ exposes this function through the alias \texttt{filter\_categorical}.

\item \textbf{Relation-connectivity filtering.}
\begin{lstlisting}[style=operatorsignature]
def filter_relation_connected(
    graph: str | Path | dict[str, Any] | InstantiatedGraph,
    entities: Sequence[GraphEntity] | EntityLookupResult | EntityFilterResult | TraversalResult,
    *,
    conditions: Sequence[
        RelationFilterCondition
    ],
) -> EntityFilterResult:
\end{lstlisting}
Keeps entities connected to specified anchors through required relation types.

\item \textbf{Numeric filtering.}
\begin{lstlisting}[style=operatorsignature]
def filter_numeric(
    entities: Sequence[GraphEntity] | EntityLookupResult | EntityFilterResult | TraversalResult,
    *,
    conditions: Sequence[
        NumericFilterCondition
    ],
) -> EntityFilterResult:
\end{lstlisting}
Applies numeric equality, inequality, threshold, or interval constraints to entity properties.

\item \textbf{Set-overlap filtering.}
\begin{lstlisting}[style=operatorsignature]
def filter_set_overlap(
    entities: Sequence[GraphEntity] | EntityLookupResult | EntityFilterResult | TraversalResult,
    *,
    conditions: Sequence[
        SetOverlapFilterCondition
    ],
) -> EntityFilterResult:
\end{lstlisting}
Filters multi-valued properties by any-match, all-match, or minimum-overlap requirements against a requested set.

\item \textbf{Aggregation.}
\begin{lstlisting}[style=operatorsignature]
def aggregate_values(
    items: Sequence[GraphEntity | ProjectedRow | dict[str, Any]] | Iterable[GraphEntity | ProjectedRow | dict[str, Any]],
    *,
    request: AggregateRequest,
) -> AggregateResult:
\end{lstlisting}
Computes count, sum, minimum, maximum, or average statistics over entities, optionally grouped by a field.

\end{itemize}

\section{Multi-Seed Robustness}
\label{app:robustness}

\begin{table}[t]
\centering
\small
\setlength{\tabcolsep}{6pt}
\renewcommand{\arraystretch}{1.12}
\resizebox{\columnwidth}{!}{
\begin{tabular}{lccccc}
\toprule
\multicolumn{1}{c}{Seed} & \multicolumn{2}{c}{CS} & \multicolumn{2}{c}{HC} & Final \\
\cmidrule(lr){2-3}\cmidrule(lr){4-5}
& Micro & Macro & Micro & Macro &  \\
\midrule
1 & 83.50 & 55.90 & 58.31 & 62.40 & 54.00 \\
2 & 88.77 & 60.40 & 60.46 & 61.90 & 58.20 \\
3 & 86.06 & 59.50 & 59.16 & 60.40 & 55.50 \\
\midrule
Avg. $\pm$ Std. & $86.11 \pm 2.64$ & $58.60 \pm 2.38$ & $59.31 \pm 1.08$ & $61.57 \pm 1.04$ & $55.90 \pm 2.13$ \\
\bottomrule
\end{tabular}
}
\caption{Multi-seed robustness of OaK on TravelPlanner using DeepSeek-v4-flash.
All entries are percentages (\%).}
\label{tab:travelplanner-robustness}
\end{table}

We report three runs of OaK with different random seeds under DeepSeek-v4-flash.
Across seeds, Table~\ref{tab:travelplanner-robustness} reports a final pass rate of $55.90 \pm 2.13$ for OaK.
The macro metrics are also stable, with $58.60 \pm 2.38$ on CS Macro and $61.57 \pm 1.04$ on HC Macro. This suggests that the gains in Table~\ref{tab:travelplanner} are not driven by a single favorable seed.

\section{Ablation Study on GPT-4o-mini}
\label{app:ablation}

We report the same three ablation variants as in the main-text ablation study, now using gpt-4o-mini as the backbone.

\begin{figure}[H]
\centering
\includegraphics[width=\columnwidth]{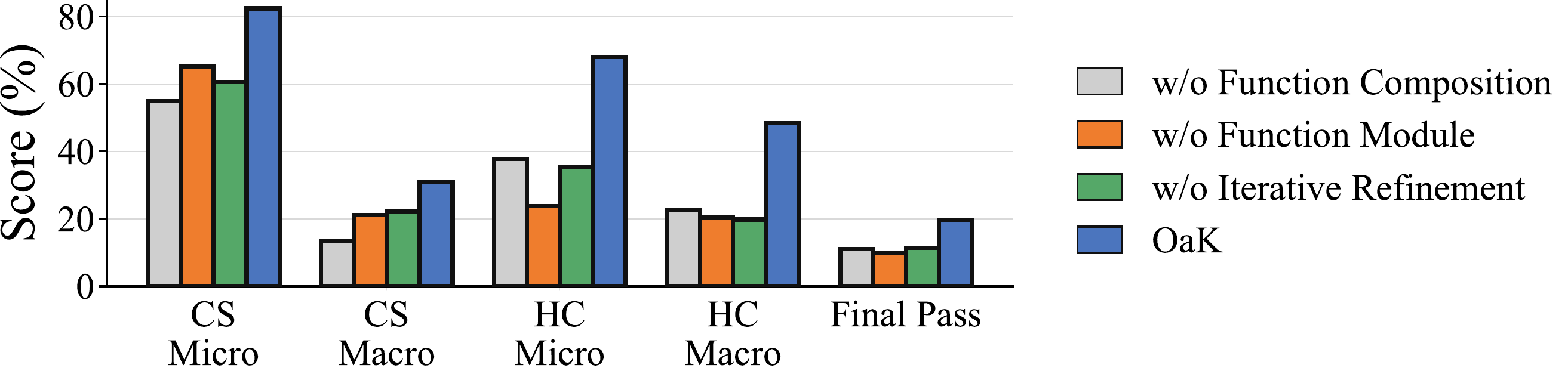}
\caption{Ablation results on TravelPlanner using gpt-4o-mini.}
\label{fig:travelplanner-ablation-gpt}
\end{figure}

\paragraph{TravelPlanner.}
Under gpt-4o-mini, Figure~\ref{fig:travelplanner-ablation-gpt} shows that OaK achieves the highest score on all five metrics.
The final pass rate drops from 19.70 to 9.90 without the function module, and the degradation is also visible on the HC metrics.

\begin{figure}[H]
\centering
\includegraphics[width=0.8\columnwidth]{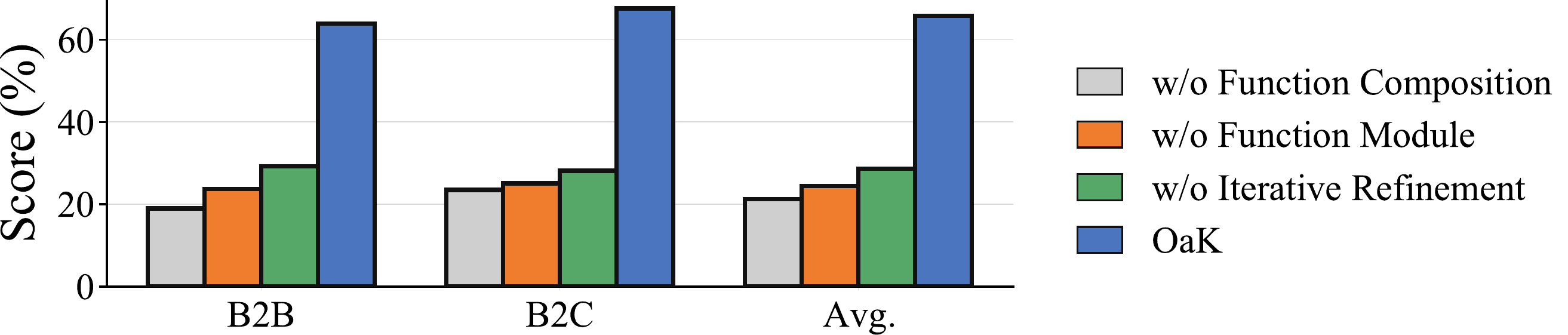}
\caption{Ablation results on CRMArenaPro using gpt-4o-mini.}
\label{fig:crmarenapro-ablation-gpt}
\end{figure}

\paragraph{CRMArenaPro.}
Figure~\ref{fig:crmarenapro-ablation-gpt} shows that the full OaK model clearly outperforms all variants on both B2B and B2C.
On the average score, OaK reaches 65.78, while the strongest ablation remains below 30 in both settings.

\begin{figure}[H]
\centering
\includegraphics[width=\columnwidth]{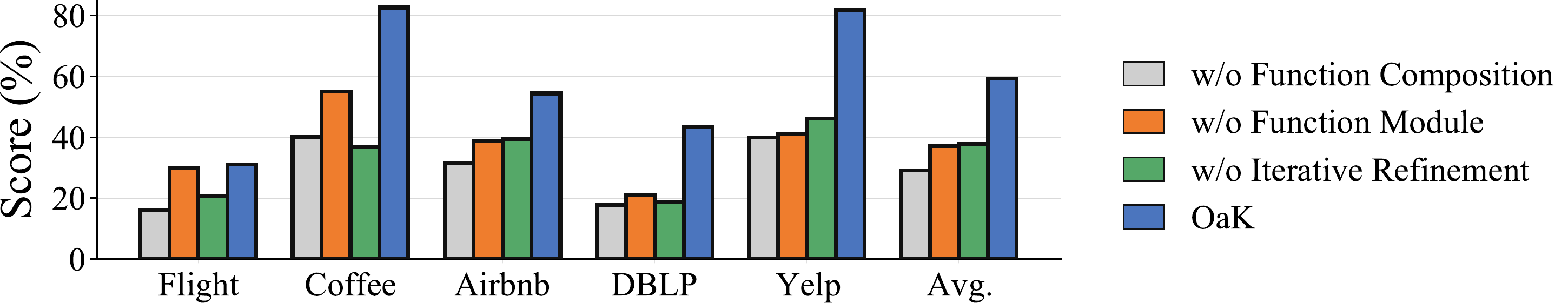}
\caption{Ablation results on ToolQA using gpt-4o-mini.}
\label{fig:toolqa-ablation-gpt}
\end{figure}

\paragraph{ToolQA.}
In Figure~\ref{fig:toolqa-ablation-gpt}, OaK outperforms all variants on every subset and reaches 59.32 on the weighted average. The ablations remain below 40.
The clearest gaps appear on Coffee, DBLP, and Yelp.
\FloatBarrier

\end{document}